\pdfoutput=1
\documentclass[10pt,twocolumn,letterpaper]{article}

\usepackage[final,algorithms]{wacv}       

\usepackage{float}

\makeatletter
\@ifpackageloaded{lineno}{%
  \setlength{\linenumbersep}{6pt}%
}{}
\makeatother

\makeatletter
\renewcommand\paragraph{\@startsection{paragraph}{4}{\z@}%
  {0.5\baselineskip \@plus 1pt \@minus 1pt}%
  {-0.5em}%
  {\normalfont\normalsize\bfseries}}
\makeatother

\makeatletter
\g@addto@macro\normalsize{%
  \setlength\abovedisplayskip{5pt plus 1pt minus 2pt}%
  \setlength\belowdisplayskip{5pt plus 1pt minus 2pt}%
  \setlength\abovedisplayshortskip{2pt plus 1pt}%
  \setlength\belowdisplayshortskip{3pt plus 1pt minus 1pt}%
}
\makeatother

\newcommand{\MechCoupOff}{-4.541}
\newcommand{\MechAbrOff}{+0.028}
\newcommand{\MechOn}{+0.364}
\newcommand{\MechOverlap}{0.00}
\newcommand{\MechOnKTwo}{+0.390}
\newcommand{\MechOnKThree}{+0.421}
\newcommand{\MechCoupOffKTwo}{-3.487}
\newcommand{\MechCoupOffKThree}{-3.214}

\newcommand{\MechCoverOn}{0.85}
\newcommand{\MechHits}{14}
\newcommand{\MechRows}{16}

\newcommand{\IrredRespLo}{74}
\newcommand{\IrredRespHi}{100}
\newcommand{\IrredSpacings}{5}
\newcommand{\IrredStepShrink}{0.12}
\newcommand{\IrredRespShrink}{0.41}
\newcommand{\IrredMiss}{0.55}
\newcommand{\IrredSteps}{6.6}
\newcommand{\IrredRefineMax}{8}
\newcommand{\IrredTrainRespLo}{74}
\newcommand{\IrredTrainSteps}{6.7}
\newcommand{\IrredTrainShrink}{0.40}
\newcommand{\CalibDiebackTrain}{0.06}
\newcommand{\HeroStemsDense}{144}
\newcommand{\HeroStemsSparse}{30}
\newcommand{\MainRel}{53.6}
\newcommand{\MainP}{0.00049}
\newcommand{\MainSeeds}{12}
\newcommand{\MainJoint}{0.415}
\newcommand{\MainIndep}{0.894}
\newcommand{\MainJointLo}{0.230}
\newcommand{\MainJointHi}{0.531}
\newcommand{\MainIndepLo}{0.882}
\newcommand{\MainIndepHi}{0.899}
\newcommand{\MainGeomRatio}{2.50}
\newcommand{\MainNoise}{0.0021}
\newcommand{\MainIso}{1.01}
\newcommand{\MainEpochs}{300}
\newcommand{\ExtrapLowRel}{62.5}
\newcommand{\ExtrapHighRel}{41.8}
\newcommand{\ExtrapP}{0.00049}

\newcommand{\TrainSpacings}{3.5, 4.5, 5.5, 6.5}
\newcommand{\CalibTarget}{0.620}
\newcommand{\CalibDieback}{0.16}
\newcommand{\CalibSeeds}{3}

\newcommand{\FieldTrees}{23}
\newcommand{\FieldMeas}{92}
\newcommand{\SpreadRatio}{1.57}
\newcommand{\ControlSpread}{31.3}
\newcommand{\Elong}{1.450}
\newcommand{\ElongLo}{1.27}
\newcommand{\ElongHi}{1.91}
\newcommand{\AnisoLattice}{1.176}
\newcommand{\AnisoBandLo}{1.20}
\newcommand{\AnisoBandHi}{1.56}
\newcommand{\AnisoTrees}{731}
\newcommand{\AnisoStands}{28}
\newcommand{\AnisoFieldMed}{1.362}

\newcommand{\JitterStar}{0.08}
\newcommand{\JitterCanonicalPct}{4}
\newcommand{\JitterStarPct}{8}
\newcommand{\JitterAniso}{1.338}
\newcommand{\JitterSpread}{1.58}
\newcommand{\JitterElong}{1.439}
\newcommand{\JitterStands}{28}
\newcommand{\HardcoreRho}{-9/10}
\newcommand{\JitterRho}{1}
\newcommand{\HardcoreOvershoot}{1.645}

\newcommand{\NSec}{24}

\newcommand{\CompGenOverlap}{9}
\newcommand{\CompIndepOverlap}{15}
\newcommand{\CompTrees}{24}
\newcommand{\GenCompSpacing}{5}
\newcommand{\GenCompTeachK}{+0.328}
\newcommand{\GenCompGenK}{+0.029}
\newcommand{\GenCompIndepK}{-0.389}
\newcommand{\DartProbeTeachK}{+0.259}
\newcommand{\DartProbeGenK}{-0.327}
\newcommand{\RealChmSpan}{96}
\newcommand{\PcSpacing}{3}
\newcommand{\PcTrees}{144}
\newcommand{\PcCoupled}{+0.180}
\newcommand{\PcUncoupled}{-7.034}
\newcommand{\PcContacted}{96}
\newcommand{\PcOtherASpacing}{5}
\newcommand{\PcOtherATrees}{51}
\newcommand{\PcOtherAContacted}{85}
\newcommand{\PcOtherBSpacing}{4}

\newcommand{\PcOtherBContacted}{92}

\definecolor{wacvblue}{rgb}{0.21,0.49,0.74}
\usepackage[pagebackref,breaklinks,colorlinks,allcolors=wacvblue]{hyperref}

\def\confName{WACV}
\def\confYear{2027}

\title{Growing a Stand, Not a Tree:\\Joint Canopy Generation Reproduces Crown Shyness}

\author{Guang Yang\\
Phi Lab Foundation\\
{\tt\small guang.yang@philab.fund}
\and
Fengchen Liu\\
University of California, Berkeley\\
{\tt\small fengchenliu@berkeley.edu}
}

\begin{document}
\maketitle
\begin{abstract}
In closed forests, neighboring tree crowns often stop short of touching, leaving a network of narrow gaps known as crown shyness. The pattern belongs to the stand rather than to any single tree, which makes it a natural probe of a question in generative modeling: can a learned model produce a set of objects whose defining structure exists only between them? We formulate stand-level canopy generation as set generation with a flow-matching model, in which attention between trees is the only channel through which coupling can arise. Trained on stands grown by a resource-competition simulation that is provably not reducible to per-tree geometry, the joint model halves the clearance distribution error of an identical-capacity model that generates each tree alone, and the advantage persists at stem densities outside the training range. Against field measurements of a tropical oak forest, a single calibrated scalar yields held-out agreement in gap magnitude and crown asymmetry. The directional statistics of the gaps are controlled by stem placement rather than by the growth rule, and match the field once stem jitter is calibrated. Crown shyness, in both the simulation and the learned model, is a property of the stand and not of the tree.
\end{abstract}

\section{Introduction}
\label{sec:intro}

Walk under a closed canopy and look up. Where two crowns meet, they often do not: a narrow channel of sky separates them, tracing the outline of each tree against its neighbors. Foresters call this \emph{crown shyness}~\cite{putz1984abrasion,vanderzee2021understanding}. The gaps are decimeters wide, they follow the shape of both crowns at once, and they vanish when either tree is removed from the picture. No single tree carries the pattern. It lives in the space between trees, maintained by the pair that shares it.

That property makes crown shyness an unusually clean test case for generative modeling of object sets. Most generative models of natural objects synthesize one instance at a time: one face, one chair, one tree~\cite{lee2024treed,khanam2026treesrnf,marcozzi2026treeflow}. When a scene needs many instances, the usual recipe places independently generated objects side by side, and any interaction between them is imposed afterward by a procedural rule. The classic ecosystem pipeline works this way: plants are modeled individually, and where two overlap, a rule deletes or subordinates the weaker one~\cite{deussen1998realistic}. It produces plausible pictures. It also guarantees, by construction, that nothing in the result depends on the objects having been generated together. If inter-object structure is what we care about, a resolution rule cannot be the answer; the structure must emerge from joint generation, or it is not being modeled at all.

This paper asks whether a learned model can generate a whole stand at once such that crown shyness emerges between its trees. The question splits into three parts, and each part carries a failure mode that has to be ruled out rather than assumed away.

\begin{figure*}[t!]
  \centering
  \includegraphics[width=\textwidth]{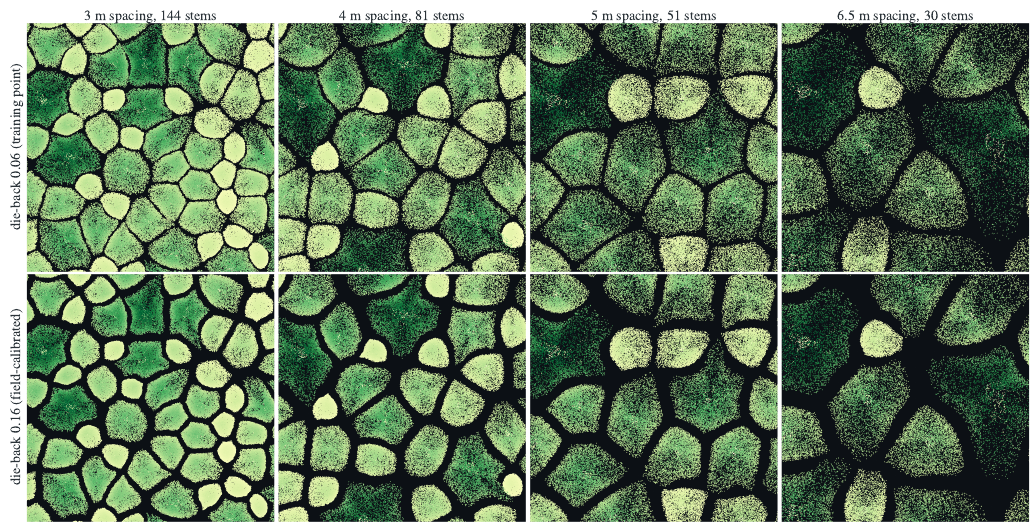}
  \caption{Crown shyness from one growth rule, across the density range this
  paper studies. Each panel is a stand grown by the coupled simulation
  (Sec.~\ref{sec:teacher}) on hard-core Poisson stems, seen from above;
  color is height, gaps are unlit background. Columns: mean stem spacing
  3.0 to 6.5~m (\HeroStemsDense{} to \HeroStemsSparse{} trees in the same
  footprint). Rows: the paper's two operating points of the contact
  die-back fraction, \CalibDiebackTrain{} selected by the mechanism gate
  and used for all training data, and \CalibDieback{} calibrated to field
  measurements (Sec.~\ref{sec:experiments}); crown draws are shared within
  each column, so the only change down a column is the abrasion strength,
  and the channels widen accordingly. The channels follow the outline of
  both crowns that share them, at every density: the signature of crown
  shyness. The learning question of this paper is whether a generative
  model can produce these stands \emph{jointly} (Sec.~\ref{sec:experiments}).}
  \label{fig:hero}
\end{figure*}

\paragraph{The training data must contain real coupling.}
If the ground truth itself reduces to a function of stem positions, a model that learns that function has learned single-tree geometry, not interaction. We therefore train on stands grown by a resource-competition simulation (Fig.~\ref{fig:hero}) and prove, before any learning, that it is not reducible: perturbing one tree's height changes its neighbors' crowns at \IrredRespLo{} to \IrredRespHi{}\% of stems across \IrredSpacings{} spacings, the response survives an \IrredRefineMax{}-fold refinement of the growth step, and the best-fitting closed-form truncation of independently grown crowns still misses by \IrredSteps{} growth steps (Sec.~\ref{sec:teacher}).

\paragraph{The comparison must isolate joint generation.}
We train two flow-matching models~\cite{lipman2023flow} of identical capacity on identical data. One attends across the trees of a stand~\cite{vaswani2017attention,lee2019set}; the other processes each tree alone. The only difference is whether trees can see each other. On stem spacings never seen in training, the joint model reaches a clearance distribution error of \MainJoint{}~m against \MainIndep{}~m for the independent model, a \MainRel{}\% improvement at $p=\MainP$ over \MainSeeds{} seeds with no overlap between the two ranges (Sec.~\ref{sec:experiments}). Shuffling stem positions within a stand degrades the joint model by \MainGeomRatio$\times$, confirming that it uses neighbor geometry rather than a memorized average. The advantage holds outside the training range, at both denser and sparser stands than the model ever saw.

\paragraph{The output must match reality, not just the simulation.}
A model can beat a baseline on simulated data and still describe no real forest. We compare against published field measurements of crown shyness in a tropical montane oak forest~\cite{markham2021wind}, calibrating exactly one scalar (median gap width) and testing on quantities the calibration never touched. Gap spread and crown asymmetry land inside the field bands; crown asymmetry in particular is a prediction, since nothing in the pipeline was ever fitted to crown shape. The directional statistics of the gaps expose a subtler lesson: on the regular stem lattice used for controlled comparisons, gap direction is too uniform, and a designed attribution experiment shows the shortfall is a property of stem placement, not of the growth rule. Calibrating stem jitter on one field scalar brings the directional statistic into the field band while leaving gap magnitude and crown shape intact (Sec.~\ref{sec:experiments}).

\paragraph{Contributions.}
(1) We formulate stand-level canopy generation as joint set generation and provide, to our knowledge, its first learned instantiation in which inter-crown structure emerges rather than being imposed. (2) We introduce an irreducibility protocol that certifies the training distribution contains genuine inter-object coupling before any model is trained on it. (3) We show with matched-capacity ablations that attention between objects, not capacity or data, is what buys the coupling, and that the advantage extrapolates. (4) We validate against field data under a one-scalar calibration with held-out tests, and we isolate which aspects of realism are controlled by the growth rule and which by stem placement.

\section{Related Work}
\label{sec:related}

\paragraph{Single-tree generative models.}
Learned generation of individual trees is an active area. Recent systems span latent developmental rules~\cite{zhou2023deeptree}, Riemannian shape spaces for static and growing trees~\cite{khanam2024riemannian,khanam2026treesrnf}, diffusion over tree point clouds and voxels~\cite{xu2025difftree,li2024svdtree}, image-conditioned reconstruction with diffusion priors~\cite{lee2024treed}, and flow matching from inventory scalars~\cite{marcozzi2026treeflow}. A survey of 585 stand- or canopy-scale titles found twenty-three learned single-tree generators and no learned model that generates a stand jointly; the four candidates that mention stand scale are a reconstruction system, an authoring tool, a procedural rule set, and an image-to-height regressor. All of these generate or process one plant at a time, and none models tree-to-tree interaction. TreeFlow is the closest in machinery, and it is not a baseline for our task: it conditions on species, platform, and height, and contains no mechanism by which one tree could influence another. We state this as a property of the field rather than a gap in our search, and we return to what can and cannot be compared in Sec.~\ref{sec:experiments}.

\paragraph{Procedural ecosystems.}
Graphics has long produced convincing forests by growing or placing plants independently and resolving interaction with explicit rules, from the ecosystem simulations of Deussen \etal~\cite{deussen1998realistic} to self-organizing tree models in which crowns compete for space during growth~\cite{palubicki2009self}. The realism of these systems is not in question, and competition-for-space growth is the direct ancestor of the simulation we train on. Their relevance here is as the boundary of the task definition: when inter-crown gaps are produced by a hand-written rule, whether carving after generation or a fixed competition procedure, the gaps encode the rule, not a learned relationship. Our aim is precisely the structure such pipelines cannot certify, coupling that emerges inside a learned generator.

\paragraph{Crown shyness in ecology.}
Crown shyness has been documented for a century. Putz \etal correlated intercrown spacing with the summed sway of adjacent crowns in still air, establishing mechanical abrasion between swaying crowns as the leading explanation~\cite{putz1984abrasion}; direct sway measurements later recorded hundreds of crown collisions per hour in moderate wind~\cite{rudnicki2002sway}. The gap pattern varies with species and crown mechanics~\cite{offermans1986shyness,onoda2021shyness} and has been analyzed in 3-D from terrestrial laser scans~\cite{vanderzee2021understanding}. Quantitative field data at the scale of individual gaps remains scarce. Markham and Fern\'andez Ot\'arola measured crown-edge clearance in four cardinal directions on \FieldTrees{} focal trees in a Costa Rican montane oak forest and released the data openly~\cite{markham2021wind}; those measurements are our field reference, including a per-tree directional statistic we test against in Sec.~\ref{sec:exp-aniso}. Our simulation models abrasion without an explicit wind field; we make no directional claim on wind's behalf, and Sec.~\ref{sec:exp-aniso} locates directional structure in stem placement instead. We note a scale distinction the paper depends on: the remote-sensing literature on canopy gaps measures treefall openings on rasters at hectare scale~\cite{asner2013gaps,hunter2015gaps}, a different quantity from the sub-meter crown-edge clearances studied here, and the two must not be conflated.

\paragraph{Set generation.}
Generating a set of interdependent objects is a general problem; permutation-invariant architectures~\cite{lee2019set} and attention~\cite{vaswani2017attention} supply the machinery, and modern continuous generators are trained by diffusion~\cite{ho2020ddpm,luo2021diffusion,zeng2022lion} or flow matching~\cite{lipman2023flow}, with point-cloud generation as the canonical instance~\cite{achlioptas2018learning,yang2019pointflow}. Closer to our setting, scene-synthesis systems generate room layouts jointly, with recent work adding physical constraints such as collision avoidance~\cite{yang2024physcene}; the objects themselves remain rigid assets whose shapes do not respond to their neighbors. In our task the objects deform each other: a crown's shape is a function of the crowns beside it. We use standard components deliberately, since the paper's claim is not architectural novelty but a controlled demonstration that, at matched capacity, the attention channel is what buys emergent inter-object structure.

\section{A Teacher That Cannot Be Faked}
\label{sec:teacher}

Learning coupling requires data that contains coupling. This section describes the simulation that generates our training stands and the protocol that certifies, before any learning, that its stands cannot be reproduced by per-tree geometry. We refer to the simulation as the \emph{teacher}.

\subsection{Coupled growth with contact abrasion}
\label{sec:teacher-rule}

A stand is a set of stems at positions $\mathbf{p}_i \in \mathbb{R}^2$ with per-tree crown parameters (height $h_i$, target radius, lobing) drawn from fixed distributions. Each crown is a radial function over \NSec{} angular sectors: tree $i$ occupies radius $r_i(\theta_s)$ in sector $s$, growing toward a lobed target
\begin{equation}
  r_i^{\ast}(\theta_s) \;=\; R_i \bigl(1 + a_i \cos(k_i \theta_s + \phi_i)\bigr),
  \label{eq:target}
\end{equation}
where $R_i$ is the open-grown radius and $(a_i, k_i, \phi_i)$ set the lobing. All crowns grow in synchronized rounds. In each round, taller trees act first and advance each unfrozen sector by a vigour-scaled step,
\begin{equation}
  r_i(\theta_s) \leftarrow r_i(\theta_s) + \Delta \cdot \bigl(h_i / \bar{h}\bigr)^{\gamma},
  \label{eq:growth}
\end{equation}
with $\gamma = 1$ throughout. Two mechanisms couple the trees. First, occupancy: a sector tip that lands inside a neighbor's current crown freezes where it is, regardless of relative height. Height asymmetry acts only through order and speed in Eq.~\eqref{eq:growth}, so a taller tree claims contested space first and the boundary settles nearer the shorter tree; no crown ever passes through another. Second, abrasion: when growth stops because of contact, the contacting sector dies back by a fixed fraction of its own length,
\begin{equation}
  r_i(\theta_s) \leftarrow (1 - \beta)\, r_i(\theta_s)
  \quad \text{where sector } s \text{ made contact},
  \label{eq:abrasion}
\end{equation}
modeling abrasion of growing tips as crowns sway, the mechanism the field literature supports~\cite{putz1984abrasion,rudnicki2002sway,markham2021wind}.

Equation~\eqref{eq:abrasion} is the point of the design: $\beta$ is dimensionless, and no parameter of the teacher is denominated in meters. Gap width is not set anywhere. It is an equilibrium between growth pressure and contact loss, so it scales with crown size and is a prediction of the model, testable against field measurements. A self-check measures worst-case crown overlap after every coupled growth and aborts on violation (overlap $\MechOverlap$~m in all runs used here).

\begin{figure*}[t!]
  \centering
  \includegraphics[width=\textwidth]{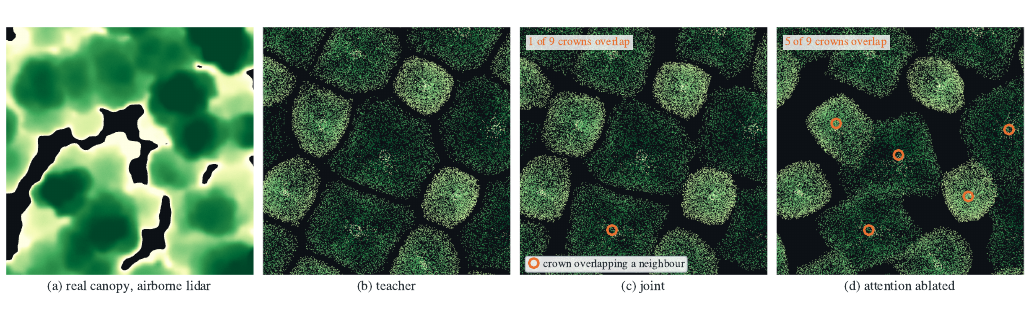}
  \caption{Crown shyness, real and generated, seen the same way: from above,
  coloured by height, gaps dark. (a) A \RealChmSpan~m window of an airborne
  lidar canopy height model, Maliau Basin, Borneo~\cite{swinfield2020lidar},
  chosen by shape search for canopy cover matching (b); interpolation for
  display only. (b)--(d) The teacher at \GenCompSpacing~m stem spacing held
  out of training (\GenCompTeachK~m clearance), the trained joint model on
  the same stems (\GenCompGenK~m), and the same model with attention ablated
  (\GenCompIndepK~m). Rings mark crowns that run into a neighbor:
  \CompGenOverlap{} of \CompTrees{} for the joint arm, \CompIndepOverlap{}
  for the ablated one. Both paint the same canopy area from the same mean
  crown radius, so they differ in crown shape, not size.}
  \label{fig:teaser}
\end{figure*}

Figure~\ref{fig:teaser} places the three canopies side by side: a real one, the teacher's, and the trained model's, all seen from above and coloured by height. Panels (b) and (c) share one set of stems at a spacing the model never saw in training; the gap network in (c) is generated, not grown, and both echo the channel structure of the lidar canopy in (a).

\subsection{Both mechanisms are necessary}
\label{sec:teacher-ablation}

\begin{figure}[tb]
  \centering
  \includegraphics[width=\linewidth]{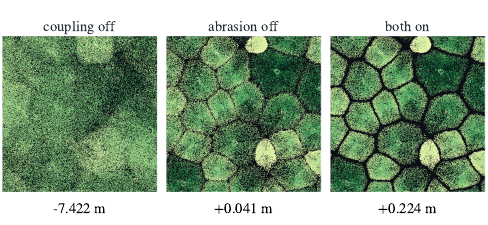}
  \caption{Shyness needs both mechanisms. Left to right: coupling off, abrasion off, both on; below each panel, nearest-neighbor clearance measured on the stand shown. Identical stems and crown draws across panels, on hard-core Poisson stems at 3.5~m; the gate itself is judged on the controlled lattice, and the rule behaves the same on both. Removing coupling interpenetrates the canopy; removing abrasion closes the gaps; only the full rule produces crown shyness.}
  \label{fig:mechanism}
\end{figure}

We ablate the two mechanisms at identical stems, crown draws, and seed (Fig.~\ref{fig:mechanism}). With coupling off, crowns interpenetrate (clearance \MechCoupOff{}~m toward the nearest neighbor, \MechCoupOffKTwo{}~m and \MechCoupOffKThree{}~m toward the second and third). With coupling on but abrasion off, crowns meet and stop, leaving no gap (\MechAbrOff{}~m). With both on, gaps open to \MechOn{}~m, \MechOnKTwo{}~m, and \MechOnKThree{}~m toward the first three neighbors. The two mechanisms are not redundant: occupancy decides \emph{where} a crown stops, abrasion decides \emph{how far back} it then retreats, and only their composition leaves a gap where two crowns met. Canopy cover moves the other way, from \MechCoverOn{} with both mechanisms to higher values without them, which is the expected trade: gaps cost cover, and a rule that maximised cover alone would close them.

The gate is deliberately strict. Canopy cover and positive clearance must hold toward the first \emph{three} nearest neighbors simultaneously, because a gate that inspects only the nearest neighbor cannot fail for a rule shaped like a nearest-neighbor rule. That error is not hypothetical; it invalidated an earlier version of this experiment, whose rule suppressed growth only across the bisector with the nearest neighbor and whose gate measured clearance only in that same direction. Measured afterwards toward the second neighbor, that stand had negative clearance. Of \MechRows{} swept configurations, \MechHits{} satisfy the gate, and a single die-back fraction covers every spacing tested, so the operating point is a regime rather than a narrow tuning.

\subsection{Irreducibility}
\label{sec:teacher-irred}

\begin{figure}[t!]
  \centering
  \includegraphics[width=\linewidth]{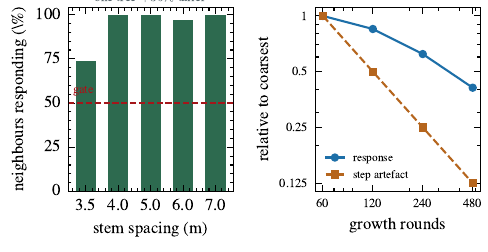}
  \caption{The teacher is not a function of stem positions. Left: fraction of neighbors whose crowns respond when one tree is made 30\% taller, by stem spacing; the dashed line marks the 50\% gate. Right: the response under step refinement (circles) against the decay a discretization artifact would show (squares, dashed); the response converges instead of tracking the step.}
  \label{fig:irreducible}
\end{figure}

The teacher would be worthless as ground truth if its stands could be written as a function of stem positions, because then a model could match them with single-tree geometry plus a lookup. The classical form of that reduction, in the tradition of ecosystem pipelines that generate plants independently and resolve interaction by rule~\cite{deussen1998realistic}, is Voronoi-style truncation: grow each tree alone, then cut its crown at the territory boundary. Formally, the reduction claims there exists a standoff $\sigma$ such that
\begin{equation}
  r_i(\theta) \;=\; \min\Bigl(r_i^{\ast}(\theta),\; b_i(\theta) - \sigma\Bigr)
  \quad \text{for all } i, \theta,
  \label{eq:reduction}
\end{equation}
where $b_i(\theta)$ is the distance from stem $i$ to the territory boundary along $\theta$, a quantity determined by stem positions alone. If Eq.~\eqref{eq:reduction} held, neighbor sizes would be irrelevant and the teacher would carry no information a per-tree model could not recover. We test it three ways (Fig.~\ref{fig:irreducible}), with all three gates fixed before the implementation.

The first test perturbs one tree, making it 30\% taller while stem positions, seeds, and every other tree are held fixed. Under Eq.~\eqref{eq:reduction} nothing else can change, since $b_i$ and $\sigma$ are untouched. Instead \IrredRespLo{} to \IrredRespHi{}\% of neighbors change shape, across \IrredSpacings{} stem spacings. The second test asks whether that response is real or an artifact of synchronized rounds: a difference that shrinks in proportion to the growth step is numerical, not mechanistic. Refining the step \IrredRefineMax$\times$ shrinks the step to \IrredStepShrink$\times$ its coarsest value while the response only shrinks to \IrredRespShrink$\times$, so it converges rather than vanishing. This control has teeth. An earlier version of the teacher failed exactly this test, its apparent coupling tracking the step size, and was retired on that evidence alone.

The third test gives the reduction its best chance. Rather than fixing $\sigma$ a priori, we sweep it and select the closed form that best fits the teacher's own output, so the sceptic's model is fitted to our data before being judged. The best fit still misses by \IrredMiss{}~m per sector, \IrredSteps{} growth steps, which is far outside anything discretization could explain. The reason is visible in the growth rule: contested space is resolved by the vigour factor in Eq.~\eqref{eq:growth}, so where a boundary settles depends on the \emph{heights} of both trees, and no function of stem coordinates can express that. Crown shape here is a property of the neighborhood, which is what makes the stands worth learning from.

\section{Method: Joint Generation of a Stand}
\label{sec:method}

\subsection{Task}
A stand is an unordered set of trees. Tree $i$ carries its stem position $\mathbf{p}_i$ and a crown vector $\mathbf{r}_i \in \mathbb{R}^{\NSec}$ of directional radii. Given stem positions and stand-level conditions $\mathbf{c}$ (spacing, local density statistics, median height), the task is to generate all crown vectors jointly so that the stand's inter-crown structure matches the teacher's. We work on patches of the $k=24$ trees nearest a focal stem, with positions normalized by spacing, so the generator sees local neighborhood geometry rather than absolute coordinates.

\subsection{Generator}
We train a conditional flow-matching model~\cite{lipman2023flow}. Each crown is decomposed into its mean radius $\bar{r}_i$, provided as an input feature, and the deviation profile $\mathbf{x}_i = \mathbf{r}_i - \bar{r}_i \mathbf{1}$, which is the generation target; the network learns crown shape given crown size, not size itself. Writing $\mathbf{x} = (\mathbf{x}_1, \ldots, \mathbf{x}_k)$ for the set and $\mathbf{f}_i = (\mathbf{p}_i / d,\, \bar{r}_i)$ for per-tree features at spacing $d$, training follows the standard conditional flow-matching objective on linear probability paths. With $\mathbf{x}_0 \sim \mathcal{N}(\mathbf{0}, \mathbf{I})$, $\mathbf{x}_1$ a data sample, and $\mathbf{x}_t = (1-t)\,\mathbf{x}_0 + t\,\mathbf{x}_1$,
\begin{equation}
  \mathcal{L}(\theta) \;=\;
  \mathbb{E}_{t,\, \mathbf{x}_0,\, \mathbf{x}_1}
  \bigl\lVert v_\theta(\mathbf{x}_t,\, t,\, \mathbf{f},\, \mathbf{c})
  - (\mathbf{x}_1 - \mathbf{x}_0) \bigr\rVert^2 ,
  \label{eq:fm}
\end{equation}
and sampling integrates the learned velocity field from noise in 100 Euler steps. The field $v_\theta$ is a stack of transformer blocks~\cite{vaswani2017attention}: each tree's noisy profile and features embed to a token, stand conditions and flow time enter through a shared embedding added to every token, and multi-head attention across the patch~\cite{lee2019set} lets trees exchange information.

Attention across trees is the only channel through which one tree's crown can influence another's. This is the experimental lever of the paper: disabling attention yields an architecture of identical capacity in which each tree is generated alone,
\begin{equation}
  v_\theta^{\text{joint}} = v_\theta \bigl(\mathbf{x}_t \mid \text{all } k \text{ trees}\bigr),
  \quad
  v_\theta^{\text{indep}} = v_\theta \bigl(\mathbf{x}_t \mid \text{tree } i\bigr).
  \label{eq:arms}
\end{equation}
We train both arms on the same data, seeds, and schedule (\MainEpochs{} epochs each) and compare them under one instrument.

\subsection{One instrument, guarded}
\label{sec:method-instrument}

The measured quantity throughout is nearest-neighbor crown-edge clearance. For trees $i$ and $j$ with stems $\nu = \lVert \mathbf{p}_j - \mathbf{p}_i \rVert$ apart,
\begin{equation}
  g(i, j) \;=\; \nu \;-\; r_i(\theta_{ij}) \;-\; r_j(\theta_{ji}),
  \label{eq:clearance}
\end{equation}
where $\theta_{ij}$ is the bearing from $i$ toward $j$; positive $g$ is a gap, negative $g$ is interpenetration. Equation~\eqref{eq:clearance} lives in one canonical estimator shared by every experiment, including the field comparisons. The evaluation error is the Wasserstein-1 distance between generated and teacher clearance distributions,
\begin{equation}
  W_1(\hat{G}, G) \;=\; \int_0^1 \bigl| \hat{F}^{-1}(u) - F^{-1}(u) \bigr|\, du,
  \label{eq:w1}
\end{equation}
with $F$ the CDF of teacher clearances and $\hat{F}$ of generated ones. Because the estimator is the paper's single instrument, the batched copy used inside the training loop asserts agreement with the canonical implementation (tolerance $10^{-6}$) before every run, and aborts on mismatch.

Three reference levels calibrate what an error value means. A noise floor (\MainNoise{}~m) from split halves of the teacher's own stands bounds attainable error from below. A direction-blind baseline (\MainIso{}~m), which replaces every crown with its circular mean, is the level a model reaches with no directional structure at all. And a shuffled-geometry probe evaluates the trained joint model with stem positions permuted inside each patch, destroying the coupling information while leaving marginals intact; a joint model that truly uses neighbor geometry must degrade.

\subsection{Preregistered gates}
Verdicts follow gates fixed before the runs. Writing $m_J$ and $m_I$ for the median $W_1$ error of the joint and independent arms over seeds, the main gate requires
\begin{equation}
  \frac{m_I - m_J}{m_I} \;\geq\; 0.20
  \qquad \text{and} \qquad
  p_{\text{Wilcoxon}} \;<\; 0.05
  \label{eq:gate}
\end{equation}
on unseen spacings over \MainSeeds{} paired seeds. Preconditions require the joint model to sit below half the direction-blind baseline (it learned structure) and to degrade by at least $1.30\times$ under shuffled geometry (it uses the stems). Twelve seeds are used because at five seeds the smallest attainable two-sided Wilcoxon $p$ is $0.0625$ and Eq.~\eqref{eq:gate} would be unsatisfiable; the repair made the test harder, not easier. At twelve seeds the smallest attainable $p$ is $2/2^{12} \approx 0.00049$, reached exactly when all twelve paired differences share one sign. The $p$ values reported in Sec.~\ref{sec:experiments} sit at this floor, so they state that every seed agreed, not how large the effect is; effect size is carried by the medians and their ranges.

\section{Experiments}
\label{sec:experiments}

\paragraph{Data and operating points.}
Stands are grown by the teacher at stem spacings 3.5 to 6.5~m on jittered lattices (rotation, phase, and 4\% positional jitter), with splits disjoint by spacing: training on \TrainSpacings~m, evaluation on the unseen spacings 4.0, 5.0, and 6.0~m, and extrapolation beyond the training range at 3.0 and 7.5~m. Coupled stands and their uncoupled twins are built from identical stems and crown draws, fixing the ablation at data level. The teacher runs at two operating points in this paper, and they serve different questions. The learning experiments (Secs.~\ref{sec:exp-main} and~\ref{sec:exp-extrap}) fix the die-back fraction at \CalibDiebackTrain{}, the value selected by the mechanism gate of Sec.~\ref{sec:teacher-ablation}, so that stem spacing is the only variable that moves. The field comparisons (Secs.~\ref{sec:exp-field} and~\ref{sec:exp-aniso}) calibrate the same parameter to field data and arrive at \CalibDieback{}; they test the teacher, not the learned model. The irreducibility certificate of Sec.~\ref{sec:teacher-irred} holds at both values: at \CalibDiebackTrain{}, the perturbation response stays at \IrredTrainRespLo{} to 100\% of neighbors, survives refinement at \IrredTrainShrink$\times$, and the best closed form still misses by \IrredTrainSteps{} growth steps. No learned baseline exists for stand-level joint generation (Sec.~\ref{sec:related}); the independent arm at matched capacity is the strongest available comparison, and the direction-blind baseline and noise floor bound the scale.

\subsection{Joint generation beats independent generation}
\label{sec:exp-main}

\begin{figure}[t!]
  \centering
  \includegraphics[width=\linewidth]{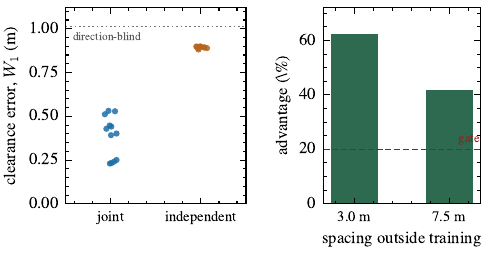}
  \caption{Joint generation beats independent generation at matched capacity. Left: clearance distribution error on unseen stem spacings for the joint and independent arms over \MainSeeds{} seeds; the dotted line is the direction-blind baseline. Right: relative advantage of joint over independent at spacings outside the training range; the dashed line is the preregistered 20\% gate.}
  \label{fig:main}
\end{figure}

Figure~\ref{fig:main} (left) shows the main result. On unseen spacings, the joint model reaches a median clearance error of \MainJoint{}~m against \MainIndep{}~m for the independent arm: a \MainRel\% relative improvement, Wilcoxon $p=\MainP$, with no overlap between the two ranges over \MainSeeds{} seeds (joint \MainJointLo{}--\MainJointHi{}, independent \MainIndepLo{}--\MainIndepHi{}). Both preconditions hold: the joint model sits far below the direction-blind baseline, and shuffling stem positions inside patches degrades it by \MainGeomRatio$\times$, confirming that the advantage flows through neighbor geometry rather than through better marginals. The independent arm's failure is not a capacity artifact; it has the same parameters, data, and schedule, and it converges to nearly seed-independent error (range \MainIndepLo{}--\MainIndepHi{}), the signature of a model that has learned everything available to it and is missing an input.

\subsection{The advantage extrapolates}
\label{sec:exp-extrap}

\begin{figure*}[t!]
  \centering
  \includegraphics[width=\textwidth]{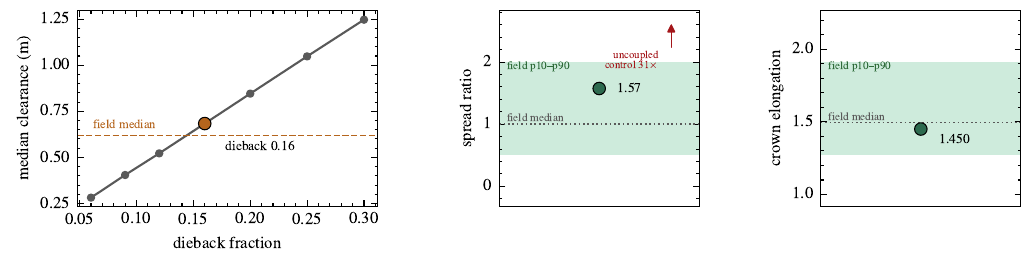}
  \caption{Field validation calibrates one scalar and holds out the rest. Left: the calibration curve; the die-back fraction is chosen where the simulated median clearance meets the field median (dashed). Center: held-out clearance spread against the preregistered band (shaded); the uncoupled control lands at \ControlSpread$\times$ the field spread and is correctly rejected. Right: held-out crown elongation against the observed field band; crown shape was never calibrated, so this is a prediction.}
  \label{fig:field}
\end{figure*}

Outside the training range the gates are judged per spacing, so a win at one end cannot mask a loss at the other (Fig.~\ref{fig:main}, right). At 3.0~m the joint model improves on independent by \ExtrapLowRel\%; at 7.5~m by \ExtrapHighRel\%; pooled $p=\ExtrapP$; and at both spacings the joint model stays below the direction-blind baseline, ruling out an advantage bought by collapsing to featureless crowns. We report and claim only the relative advantage. Absolute extrapolation error is lower than interpolation error in our setup, but that reflects task difficulty (the 3.0~m clearance distribution is narrow, so $W_1$ is small by construction), not stronger generalization.

\begin{figure*}[t!]
  \centering
  \includegraphics[width=\textwidth]{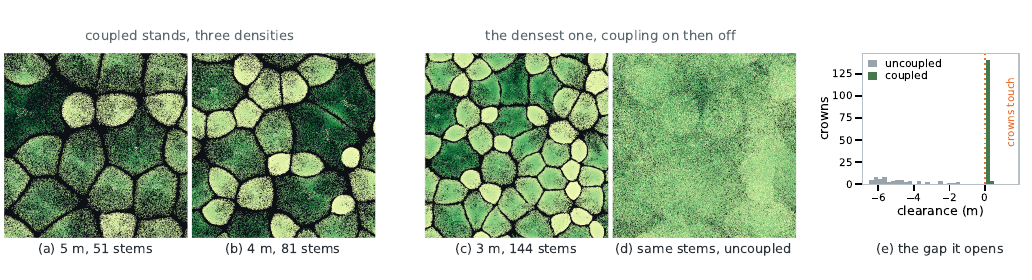}
  \caption{The gap network across stem densities, drawn as the point cloud a
  scan would return. (a--c) Coupled stands at \PcOtherASpacing, \PcOtherBSpacing{}
  and \PcSpacing~m mean stem spacing (\PcOtherATrees{} to \PcTrees{} stems),
  with \PcOtherAContacted, \PcOtherBContacted{} and \PcContacted\% of crowns
  contacting a neighbour. Stems are hard-core Poisson draws; the controlled
  experiments (Fig.~\ref{fig:main}) keep the jittered lattice. The channels
  are an artefact of neither. (d) The stand in (c) regrown with coupling
  off, from identical stems and crown draws: the canopy closes. (e)
  Nearest-neighbour clearance for (c) and (d), moving from \PcUncoupled~m
  to \PcCoupled~m.}
  \label{fig:pointcloud-stand}
\end{figure*}

\subsection{Field validation with one calibrated scalar}
\label{sec:exp-field}

Simulated gaps could be internally consistent and still describe no real forest, so we compare the teacher against field measurements of crown shyness in a tropical montane oak forest: \FieldTrees{} focal trees, \FieldMeas{} directional crown-edge clearance measurements, published openly~\cite{markham2021wind}. The teacher's single free parameter, the die-back fraction lost on contact, is calibrated to one scalar, the field tree-mean median clearance of \CalibTarget{}~m, on \CalibSeeds{} seeds. The calibration selects \CalibDieback{} (Fig.~\ref{fig:field}, left). This value is independent of the training operating point of Sec.~\ref{sec:exp-main}; the two are separate settings of the same dial, each answering its own question. Every test below runs on disjoint seeds and quantities the calibration never touched.

Held out, the clearance distribution spread lands at \SpreadRatio$\times$ the field spread, inside the preregistered band of 0.5 to 2.0 (Fig.~\ref{fig:field}, center). Crown elongation, the ratio of the longest to shortest crown axis, lands at \Elong{} inside the observed band of \ElongLo{} to \ElongHi{} (Fig.~\ref{fig:field}, right). Elongation deserves emphasis: nothing in the teacher or the model was ever calibrated to crown shape, so this agreement is a prediction. The comparison also keeps its power after calibration: the uncoupled twin stand is rejected by the same spread gate at \ControlSpread$\times$ the field value, so the test still separates the mechanism from its absence.

\begin{figure}[t!]
  \centering
  \includegraphics[width=\linewidth]{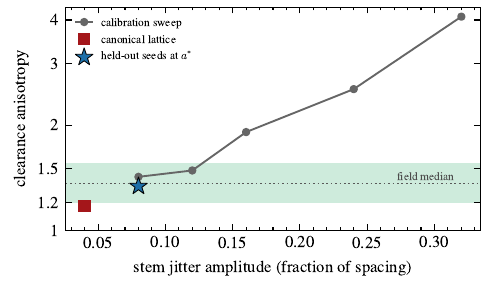}
  \caption{Directional gap statistics are controlled by stem placement. Clearance anisotropy against stem jitter amplitude (circles: calibration sweep; square: the controlled lattice used in the learning experiments; star: held-out evaluation at the calibrated amplitude). The shaded band is the field's observed range and the dotted line its median. Raising jitter from \JitterCanonicalPct\% to \JitterStarPct\% moves anisotropy into the field band; gap magnitude and crown shape remain inside their own bands.}
  \label{fig:anisotropy}
\end{figure}

\subsection{Gap direction is controlled by stem placement}
\label{sec:exp-aniso}

The field data reports one more statistic: per-tree anisotropy, the ratio of a tree's widest to narrowest cardinal gap, with median \AnisoFieldMed{} and observed band \AnisoBandLo{} to \AnisoBandHi{}~\cite{markham2021wind}. On the controlled lattice the model measures \AnisoLattice{} across \AnisoTrees{} trees in \AnisoStands{} stands: below the band. The gaps are too uniform in direction. This verdict survived three successive corrections of the measurement itself (quantity, reference frame, estimator provenance), so it is a property of the pipeline, not an artifact.

Which part of the pipeline? The controlled lattice carries \JitterCanonicalPct\% stem jitter, chosen so that spacing could act as a controlled variable in the learning experiments, never as a claim about real stands. On such a lattice the stem geometry alone caps directional variation: the four cardinal neighbor distances are nearly equal by construction. We therefore ran a designed attribution experiment with one knob, the stem process, and everything else pinned (teacher, die-back, estimator, seeds discipline). Two stem families were tested. A hard-core family (minimum stem distance) confirms the mechanism, anisotropy falling monotonically as regularity rises (exact Spearman $\rho=\HardcoreRho$), but its tightest buildable level overshoots the band (\HardcoreOvershoot{}), bracketing without landing. A jitter-amplitude family interpolates outward from the lattice itself ($\rho=\JitterRho$). Calibrating the amplitude on one field scalar, the anisotropy median, selects $a^{*}=\JitterStar$; on held-out seeds this lands the anisotropy at \JitterAniso{}, inside the field band, while the spread ratio (\JitterSpread) and elongation (\JitterElong) stay inside their bands (Fig.~\ref{fig:anisotropy}). All \JitterStands{} stands built successfully.

The lattice verdict stands, and so does the lattice: the learning experiments require spacing as a controlled variable, and their conclusions are about coupling, not about directional realism. What the attribution settles is that the anisotropy shortfall belongs to stem placement, not to the growth rule, and that no ingredient of realism was sacrificed to gain it. Directional statistics on arbitrary stem patterns remain unverified; we make no claim beyond the calibrated amplitude.

\subsection{What is and is not compared}
\label{sec:exp-scope}

Before the comparisons, Fig.~\ref{fig:pointcloud-stand} states the phenomenon at the resolution a reader can judge by eye: the same stems and the same crown parameters, grown independently and then jointly, rendered as the point cloud a scan would return. The gap network in the joint panel is what every number below measures.

No published learned model generates a stand jointly (Sec.~\ref{sec:related}), so there is no external baseline to run. The closest system, TreeFlow~\cite{marcozzi2026treeflow}, generates one tree from inventory scalars; its checkpoints are unavailable and retraining is estimated at 1{,}500 GPU-hours, and even reproduced it could not express inter-tree structure. We therefore bound our result from inside: matched-capacity ablation (the independent arm), an information ablation (shuffled geometry), a structure-free baseline (direction-blind), and a measurement floor (split-half noise). Each isolates one ingredient, and together they leave joint generation as the only unexplained difference.

\section{Conclusion}
\label{sec:conclusion}

Crown shyness is a pattern that no single tree owns. We asked whether a learned generator can produce it the way forests do, between objects generated together, and the answer is yes, under conditions this paper makes precise. The training distribution must contain coupling that cannot be rewritten as per-tree geometry; we certified this with a perturbation test, a discretization control, and a best-case reduction that still misses by \IrredSteps{} growth steps. The generator must be compared against itself without the coupling channel; at matched capacity, attention between trees halves the clearance error on unseen stem densities and keeps its advantage outside the training range. And realism must be tested against measurements, not intuition; one calibrated scalar yields held-out agreement in gap magnitude and crown shape, and a designed attribution experiment locates directional realism in stem placement rather than in the growth rule.

\paragraph{Limitations.}
Our stands are two-and-a-half dimensional: each crown is a radial profile over \NSec{} sectors, not a full 3-D crown volume. The model is trained on jittered-lattice stems, and it does not survive leaving them: probed on hard-core Poisson stems, the joint arm's clearance falls to \DartProbeGenK~m against the teacher's \DartProbeTeachK~m on the same stand, so every model panel in this paper shows lattice stems, and generalizing across stem processes is open. The field reference is a single forest with \FieldTrees{} focal trees; richer directional claims are bounded by what that sample can decide, and stand-level directional structure (as opposed to per-tree anisotropy) is not decidable at this sample size. Directional statistics are verified at the calibrated jitter amplitude only. Extending the generator to full 3-D crowns and testing against modern forest laser-scanning benchmarks~\cite{puliti2023forinstance,henrich2024treelearn} are the natural next steps.

\paragraph{Reproducibility.}
Every measured statistic in this paper is generated from frozen result files by the same scripts that render the figures; none is transcribed by hand. Design constants, preregistered thresholds, and literature counts appear in the text directly, and each traces to a script, a gate definition, or a cited source. The simulation, training code, evaluation estimator, and result files will be released, together with the audit tooling that pins the estimator, the teacher version, and the gates.

{
    \small
    \bibliographystyle{ieeenat_fullname}
    \bibliography{main}

@article{marcozzi2026treeflow,
  author  = {Marcozzi, Anthony and Tenny, Johnathan and Martin, Daithi and Castorena, Juan and Crennen, Zachary and Wells, Lucas and Hillman, Samuel},
  title   = {{TreeFlow}: Conditional Flow Matching for {3D} Tree Point Cloud Generation from Inventory Attributes},
  journal = {Remote Sensing},
  volume  = {18},
  number  = {13},
  pages   = {2197},
  year    = {2026}
}

@inproceedings{lipman2023flow,
  author    = {Lipman, Yaron and Chen, Ricky T. Q. and Ben-Hamu, Heli and Nickel, Maximilian and Le, Matt},
  title     = {Flow Matching for Generative Modeling},
  booktitle = {International Conference on Learning Representations},
  year      = {2023},
  note      = {arXiv:2210.02747}
}

@inproceedings{vaswani2017attention,
  author    = {Vaswani, Ashish and Shazeer, Noam and Parmar, Niki and Uszkoreit, Jakob and Jones, Llion and Gomez, Aidan N. and Kaiser, Lukasz and Polosukhin, Illia},
  title     = {Attention Is All You Need},
  booktitle = {Advances in Neural Information Processing Systems},
  year      = {2017}
}

@inproceedings{lee2019set,
  author    = {Lee, Juho and Lee, Yoonho and Kim, Jungtaek and Kosiorek, Adam R. and Choi, Seungjin and Teh, Yee Whye},
  title     = {Set Transformer: A Framework for Attention-Based Permutation-Invariant Neural Networks},
  booktitle = {International Conference on Machine Learning},
  year      = {2019}
}

@inproceedings{ho2020ddpm,
  author    = {Ho, Jonathan and Jain, Ajay and Abbeel, Pieter},
  title     = {Denoising Diffusion Probabilistic Models},
  booktitle = {Advances in Neural Information Processing Systems},
  year      = {2020}
}

@inproceedings{lee2024treed,
  author    = {Lee, Jae Joong and Li, Bosheng and Beery, Sara and Huang, Jonathan and Fei, Songlin and Yeh, Raymond A. and Benes, Bedrich},
  title     = {Tree-D Fusion: Simulation-Ready Tree Dataset from Single Images with Diffusion Priors},
  booktitle = {European Conference on Computer Vision},
  year      = {2024}
}

@inproceedings{deussen1998realistic,
  author    = {Deussen, Oliver and Hanrahan, Pat and Lintermann, Bernd and M{\v{e}}ch, Radom{\'i}r and Pharr, Matt and Prusinkiewicz, Przemyslaw},
  title     = {Realistic Modeling and Rendering of Plant Ecosystems},
  booktitle = {SIGGRAPH},
  year      = {1998}
}

@article{vanderzee2021understanding,
  author  = {van der Zee, Jens and Lau, Alvaro and Shenkin, Alexander},
  title   = {Understanding Crown Shyness from a 3-D Perspective},
  journal = {Annals of Botany},
  volume  = {128},
  number  = {6},
  pages   = {725--735},
  year    = {2021},
  note    = {DOI 10.1093/aob/mcab035}
}

@article{markham2021wind,
  author  = {Markham, John and Fern{\'a}ndez Ot{\'a}rola, Mauricio},
  title   = {Wind Creates Crown Shyness, Asymmetry, and Orientation in a Tropical Montane Oak Forest},
  journal = {Biotropica},
  volume  = {52},
  number  = {6},
  pages   = {1127--1130},
  year    = {2020},
  note    = {DOI 10.1111/btp.12877; data: Dryad 10.5061/dryad.34tmpg4hk, CC0}
}

@article{putz1984abrasion,
  author  = {Putz, Francis E. and Parker, Geoffrey G. and Archibald, Ruth M.},
  title   = {Mechanical Abrasion and Intercrown Spacing},
  journal = {American Midland Naturalist},
  volume  = {112},
  number  = {1},
  pages   = {24--28},
  year    = {1984}
}

@techreport{rudnicki2002sway,
  author      = {Rudnicki, Mark and Lieffers, Victor J. and Silins, Uldis},
  title       = {Wind, Tree Sway and Crown Shyness in Lodgepole Pine},
  institution = {Centre for Enhanced Forest Management, University of Alberta},
  number      = {EFM Research Note 03/2002},
  year        = {2002}
}

@article{offermans1986shyness,
  author  = {Offermans, D. M. J.},
  title   = {Crown Shyness: A Parameter for Ageing in \emph{Piptadeniastrum africanum}},
  journal = {Netherlands Journal of Agricultural Science},
  volume  = {34},
  number  = {4},
  pages   = {493--497},
  year    = {1986}
}

@article{onoda2021shyness,
  author  = {Onoda, Yusuke and Bando, Erika},
  title   = {Wider Crown Shyness Between Broad-Leaved Tree Species Than Between Coniferous Tree Species in a Mixed Forest of \emph{Castanopsis cuspidata} and \emph{Chamaecyparis obtusa}},
  journal = {Ecological Research},
  volume  = {36},
  number  = {4},
  pages   = {733--743},
  year    = {2021},
  note    = {DOI 10.1111/1440-1703.12233}
}

@article{asner2013gaps,
  author  = {Asner, Gregory P. and Kellner, James R. and Kennedy-Bowdoin, Ty and Knapp, David E. and Anderson, Christopher and Martin, Roberta E.},
  title   = {Forest Canopy Gap Distributions in the Southern {P}eruvian {A}mazon},
  journal = {PLoS ONE},
  volume  = {8},
  number  = {4},
  pages   = {e60875},
  year    = {2013}
}

@article{hunter2015gaps,
  author  = {Hunter, Maria O. and Keller, Michael and Morton, Douglas and Cook, Bruce and Lefsky, Michael and Ducey, Mark and Saleska, Scott and de Oliveira, Raimundo Cosme and Schietti, Juliana},
  title   = {Structural Dynamics of Tropical Moist Forest Gaps},
  journal = {PLoS ONE},
  volume  = {10},
  number  = {7},
  pages   = {e0132144},
  year    = {2015}
}

@inproceedings{palubicki2009self,
  author    = {Pa{\l}ubicki, Wojciech and Horel, Kipp and Longay, Steven and Runions, Adam and Lane, Brendan and M{\v{e}}ch, Radom{\'i}r and Prusinkiewicz, Przemyslaw},
  title     = {Self-Organizing Tree Models for Image Synthesis},
  booktitle = {SIGGRAPH},
  year      = {2009}
}

@inproceedings{khanam2026treesrnf,
  author    = {Khanam, Tahmina and Laga, Hamid and Bennamoun, Mohammed and Wang, Guanjin and Sohel, Ferdous and Boussaid, Farid and Srivastava, Anuj},
  title     = {{TreeSRNF}: Square-Root Normal Fields for Generative Modelling of the Geometric and Structural Variability in Tree-Like {3D} Objects},
  booktitle = {European Conference on Computer Vision},
  year      = {2026},
  note      = {arXiv:2607.13456}
}

@inproceedings{khanam2024riemannian,
  author    = {Khanam, Tahmina and Laga, Hamid and Bennamoun, Mohammed and Wang, Guanjin and Sohel, Ferdous and Boussaid, Farid and Wang, Guan and Srivastava, Anuj},
  title     = {A {R}iemannian Approach for Spatiotemporal Analysis and Generation of {4D} Tree-Shaped Structures},
  booktitle = {European Conference on Computer Vision},
  year      = {2024}
}

@article{zhou2023deeptree,
  author  = {Zhou, Xiaochen and Li, Bosheng and Bene{\v{s}}, Bedrich and Fei, Songlin and Pirk, S{\"o}ren},
  title   = {{DeepTree}: Modeling Trees with Situated Latents},
  journal = {IEEE Transactions on Visualization and Computer Graphics},
  volume  = {30},
  number  = {8},
  pages   = {5795--5809},
  year    = {2024},
  note    = {arXiv:2305.05153}
}

@article{xu2025difftree,
  author  = {Xu, Haifeng and Huai, Yongjian and Nie, Xiaoying and Meng, Qingkuo and Zhao, Xun and Pei, Xuanda and Lu, Hao},
  title   = {Diff-Tree: A Diffusion Model for Diversified Tree Point Cloud Generation with High Realism},
  journal = {Remote Sensing},
  volume  = {17},
  number  = {5},
  pages   = {923},
  year    = {2025}
}

@inproceedings{achlioptas2018learning,
  author    = {Achlioptas, Panos and Diamanti, Olga and Mitliagkas, Ioannis and Guibas, Leonidas},
  title     = {Learning Representations and Generative Models for {3D} Point Clouds},
  booktitle = {International Conference on Machine Learning},
  year      = {2018}
}

@inproceedings{yang2019pointflow,
  author    = {Yang, Guandao and Huang, Xun and Hao, Zekun and Liu, Ming-Yu and Belongie, Serge and Hariharan, Bharath},
  title     = {{PointFlow}: {3D} Point Cloud Generation with Continuous Normalizing Flows},
  booktitle = {IEEE/CVF International Conference on Computer Vision},
  year      = {2019}
}

@inproceedings{luo2021diffusion,
  author    = {Luo, Shitong and Hu, Wei},
  title     = {Diffusion Probabilistic Models for {3D} Point Cloud Generation},
  booktitle = {IEEE/CVF Conference on Computer Vision and Pattern Recognition},
  year      = {2021}
}

@inproceedings{zeng2022lion,
  author    = {Zeng, Xiaohui and Vahdat, Arash and Williams, Francis and Gojcic, Zan and Litany, Or and Fidler, Sanja and Kreis, Karsten},
  title     = {{LION}: Latent Point Diffusion Models for {3D} Shape Generation},
  booktitle = {Advances in Neural Information Processing Systems},
  year      = {2022}
}

@inproceedings{li2024svdtree,
  author    = {Li, Yuan and Liu, Zhihao and Benes, Bedrich and Zhang, Xiaopeng and Guo, Jianwei},
  title     = {{SVDTree}: Semantic Voxel Diffusion for Single Image Tree Reconstruction},
  booktitle = {IEEE/CVF Conference on Computer Vision and Pattern Recognition},
  year      = {2024}
}

@inproceedings{yang2024physcene,
  author    = {Yang, Yandan and Jia, Baoxiong and Zhi, Peiyuan and Huang, Siyuan},
  title     = {{PhyScene}: Physically Interactable {3D} Scene Synthesis for Embodied {AI}},
  booktitle = {IEEE/CVF Conference on Computer Vision and Pattern Recognition},
  year      = {2024}
}

@article{henrich2024treelearn,
  author  = {Henrich, Jonathan and van Delden, Jan and Seidel, Dominik and Kneib, Thomas and Ecker, Alexander S.},
  title   = {{TreeLearn}: A Deep Learning Method for Segmenting Individual Trees from Ground-Based {LiDAR} Forest Point Clouds},
  journal = {Ecological Informatics},
  volume  = {84},
  pages   = {102888},
  year    = {2024},
  note    = {arXiv:2309.08471}
}

@article{puliti2023forinstance,
  author  = {Puliti, Stefano and Pearse, Grant and Surov{\'y}, Peter and Wallace, Luke and Hollaus, Markus and Wielgosz, Maciej and Astrup, Rasmus},
  title   = {{FOR-instance}: A {UAV} Laser Scanning Benchmark Dataset for Semantic and Instance Segmentation of Individual Trees},
  journal = {arXiv:2309.01279},
  year    = {2023}
}

@misc{swinfield2020lidar,
  author       = {Swinfield, Tom and Milodowski, David and Jucker, Tommaso and
                  Dalponte, Michele and Coomes, David},
  title        = {{LiDAR} Canopy Structure 2014 [Dataset]},
  howpublished = {Zenodo},
  year         = {2020},
  note         = {doi:10.5281/zenodo.4020697, CC-BY-4.0. Collected under the
                  NERC Human Modified Tropical Forests programme
                  (NE/K016377/1), SAFE Project}
}
}

\end{document}